\documentclass[conference]{IEEEtran}
\usepackage{cite}
\usepackage{amsmath,amssymb,amsfonts}
\usepackage{algorithmic}
\usepackage{graphicx}
\usepackage{textcomp}
\usepackage{xcolor}
\usepackage{bibentry}
\usepackage{amsfonts}
\usepackage{booktabs}
\usepackage{siunitx}
\usepackage{atbegshi}% http://ctan.org/pkg/atbegshi

\begin{document}

\title{Deep Similarity Learning for Sports Team Ranking}

\author{\IEEEauthorblockN{Daniel Yazbek, Jonathan Sandile Sibindi, Terence L. Van Zyl$^\dagger$}
\IEEEauthorblockA{\textit{Institute for Intelligent Systems} \\
\textit{University of Johannesburg}\\
Johannesburg, South Africa \\
$^\dagger$tvanzyl@gmail.com }
% \and
% \IEEEauthorblockN{Terence L. Van Zyl}
% \IEEEauthorblockA{\textit{Institute for Intelligent Systems} \\
% \textit{University of Johannesburg}\\
% Johannesburg, South Africa \\
% tvanzyl@uj.ac.za}
}
\IEEEoverridecommandlockouts
\IEEEpubid{\makebox[\columnwidth]{978-0-7381-1236-7/21/\$31.00~
\copyright2021
IEEE \hfill} \hspace{\columnsep}\makebox[\columnwidth]{ }}
\maketitle

\begin{abstract}

Sports data is more readily available and consequently, there has been an increase in the amount of sports analysis, predictions and rankings in the literature. Sports are unique in their respective stochastic nature, making analysis, and accurate predictions valuable to those involved in the sport.
In response, we focus on Siamese Neural Networks (SNN) in unison with LightGBM and XGBoost models, to predict the importance of matches and to rank teams in Rugby and Basketball.

Six models were developed and compared, a LightGBM, a XGBoost, a LightGBM (Contrastive Loss), LightGBM (Triplet Loss), a XGBoost (Contrastive Loss), XGBoost (Triplet Loss).

The models that utilise a Triplet loss function perform better than those using Contrastive loss. It is clear LightGBM (Triplet loss) is the most effective model in ranking the NBA, producing a state of the art (SOTA) mAP (0.867) and NDCG (0.98) respectively. The SNN (Triplet loss) most effectively predicted the Super 15 Rugby, yielding the SOTA mAP (0.921), NDCG (0.983), and $r_s$ (0.793). Triplet loss produces the best overall results displaying the value of learning representations/embeddings for prediction and ranking of sports. Overall there is not a single consistent best performing model across the two sports indicating that other Ranking models should be considered in the future.

\end{abstract}

\section{Introduction}
The goal of this research is to compare evaluate the extent to which deep learning of representations can improve model performance across predictions in sport. The sports considered are the National Basketball Association (NBA) teams’ regular-season performances and the Super 15 rugby. There are thirty teams in the NBA divided into two conferences - the Western Conference and the Eastern Conference. All thirty teams play 82 games in the regular season, where the top eight teams from each conference advance to the playoffs, meaning 16 out of the 30 teams make the playoffs each year~\cite{inproceedings3}. The Super 15 Competition consists of fifteen rugby teams from five countries across four continents~\cite{SuperRugRules}. Each squad plays a total of sixteen games before entering the knock-out stages.

Sports outcome prediction continues to gain popularity~\cite{manack2020deep}, as demonstrated by the globalization of sports betting~\cite{article1}. The NBA and Super 15 Rugby are some of the world's most-watched sports~\cite{inproceedings,SuperRugRules}. This popularity leaves a large number of fans anticipating the results of upcoming games. Furthermore, there are numerous betting companies offering gambling odds of one team winning over another. With the availability of machine learning algorithms, forecasting the outcome of games with high precision is becoming more feasible and economically desirable to various players in the betting industry. Further, accurate game predictions can aid sports teams in making adjustments to teams to achieve the best possible future results for that franchise. This increase in economic upside has increased research and analysis in sports prediction~\cite{article1}, using a multitude of machine learning algorithms~\cite{Lieder2018}.

% Basketball has a relatively high number of data-driven studies compared to rugby which is largely unexplored. This may be attributed to a large amount of well-documented data available to the public, on websites like Kaggle~\cite{thompson} - a well-respected data science server, dating back to as early as the 1950s. The data available includes features ranging from a player's age that season to the minutes per game he averaged. The proposed predictions of all three sports will likely be influenced by these current studies in NBA sports prediction due to how much more is readily available. With this said, there will be multiple common features across the three sports predictions, such as whether a team is playing a Home or an Away game~\cite{Ahmadalinezhad2019519}. There will be more features that differ than not due to the inherent differences in each sport, and all will struggle to control unpredictable variables such as trades/transfers and injuries.

Popular machine learning algorithms are neural networks, designed to learn patterns and effectively classify data into identifiable classes. Classification tasks depend upon labelled datasets; for a neural network to learn the correlation between labels and data, humans must transfer their knowledge to a dataset. Siamese Neural Networks (SNNs) are a type of neural network architecture. Instead of the model learning to classify its inputs, this type of neural network learns to measure the similarity between it's inputs~\cite{dlamini2019author,van2020unique}. The main idea of the Siamese network is to create two feed-forward networks with shared weights. A feed-forward network defines a mapping as $y = f(x;\theta)$ and learns the value of the parameters $\theta$ that results in the best function approximation~\cite{Goodfellow-et-al-2016}. The feed-forward-network is used to construct a function to calculate the similarity or distance metric between the two instances \cite{10.5555/2987189.2987282}. In this research, a SNN is implemented to learn a feature representation from a carefully selected group of NBA and Rugby statistics, which is then input into a ranking algorithm.

A variety of machine learning algorithms can be used to link a feature representation to a 'win' or 'loss' output. Linear regression has been used in multiple sports prediction papers~\cite{article, Torres2013}. However, there are drawbacks to this model as it fails to fit multivariate relationships~\cite{Barlos2015}. For example, a linear regression model in the NBA would likely be able to fit a relationship between 'minutes per game' (MPG) and 'points per game' (PPG) but wouldn't find success between MPG and 'rebounds per game' (RPG)~\cite{Arnold2012}. This is probably because a player's ability to get rebounds is heavily based on their position in their team, whereas scoring points are usually spread throughout the whole team and not as role-specific as rebounding. To overcome this drawback, logistic regression could be used as Lin, Short, and Sundaresan (2014)~\cite{Lin2014} suggest - in combination with other methods such as Adaptive Boost, Random Forest and SVM.

Adaptive Boost and Random Forests seek to classify games as wins or losses by averaging the results of many weaker classifiers, such as naive Bayes~\cite{Lin2014}. Lin 2014~\cite{Lin2014}, used box score data from 1991-1998 to develop machine learning models to predict the winner of professional basketball games. The methods included an adaptive boosting algorithm, random-forest model, logistic regression and a support vector machines (SVMs)~\cite{10.1145/130385.130401}. The adaptive boost algorithm is most closely related to this work and performs multiple iterations that improve a classifier by attempting to correctly classify data points that were incorrectly classified on the previous iteration. This method yielded a testing accuracy of 64.1\% in Lin 2014~\cite{Lin2014}. However, the high number of iterations performed makes the model sensitive to outliers as it will continually try predicting them correctly~\cite{Lin2014} leading, to over-fitting of data in some cases. Logistic regression yielded 64.7\% testing accuracy and the random-forest model yielded 65\% testing accuracy, just behind the SVM model at 65.2\% testing accuracy, leaving little to split the four models respective performances~\cite{Lin2014}.

In this work we will use feature learning in the form of similarity learning to learn an embedding for each of the sports considered. The learned features will be ranked using gradient boosting with the LightGBM and XGBoost implementations. The outputs are to predict the seasonal standing for the NBA and Super 15. Multiple seasons will be trained to predict a selected test season to be evaluated to gauge the performance of the models.  The use of an SNN together with a LightGBM and XGBooost for this task is an unexplored method in the current literature.

The contributions of this paper are:
\begin{enumerate}
    \item Showing that a general framework including gradient boosting and deep representation learning can successfully predict seasonal rankings across different sports (NBA and Rugby).
    \item Demonstrating that the the use of representation learning improves rankings using both the Gradient Boost Machines (LightGBM and XGBoost).
    \item Improving on SOTA in NBA and establishing the SOTA in Super 15 ranking task.
\end{enumerate}

\section{Methodology}
\subsection{Overview}
% In this study, all thirty NBA teams in the 2017/18 season were ranked using two methods. First, using a Light Gradient Boost Machine (LightGBM) \cite{Ke2017}, and secondly using a Siamese Neural Network (SNN) \cite{Goodfellow-et-al-2016}, which was further split using loss functions, Contrastive loss and Triplet loss creating a total of three models. In the second method, the resulting scores were then ranked using the LightGBM. Thereafter, the performances of our three models were compared using the following metrics: Precision, Recall, F1 score, Mean Average Precision (mAP), Normal Discounted Cumulative Gain (NDCG) \cite{inproceedings1}, and the Spearman's rank correlation coefficient \cite{EllisandVictoria2011}. Two baseline models are created, one randomly and the other naively in order to gauge the performance and usefulness of the three models.
The descriptions, preprocessing, and analysis used during the experimental process are explained in this section. The processed data is transferred to the models after partitioning. The data is partitioned by placing them into their respective seasons, whereby the first three seasons will make up the training data and the final season making up the testing data. The models are trained and tested with their results compared and analysed using appropriate metrics.

\subsection{Data}
\begin{table*}[htb!]
  \centering
  \caption{Features of Rugby team seasonal statistics dataset}
  \label{tab:1}
  \begin{tabular}{|c l | c l |c l|}
  \hline 
    1)  & Team name & 2) & Total tries scored & 3) & Home tries scored \\
    4)  & Away tries scored & 5) & Tries scored in first half & 6) & Tries scored in second half\\
    7)  & Total matches played & 8) & Tries conceded & 9)  & Tries conceded at home \\
    10) & Tries conceded whilst away & 11) & Four try bonus point & 12) & Four try bonus point at home\\
    13) & Four try bonus point whilst away & 14) & Lost within seven points & 15) &  Lost within seven points at home \\
    16) & Lost within seven points whilst away & 17) & Opponent four try bonus point & 18) & Opponent four try bonus point at home \\
    19) & Opponent four try bonus point away & 20) & Opponent lost within seven points & 21) & Opponent lost within seven points at home \\
    22) & Opponent lost within seven points away & 23) & Yellow cards accumulated & 24) & Yellow cards to red cards acquired\\
    25) & Red cards accumulated & 26) & Halftime wins & 27) & Halftime wins to full time wins\\
    28) & Halftime draws & 29) & Halftime draws to full time wins & 30) & Half time lose\\
    31) & Half time lose to full time win & 32) & Conversions &33) &  Penalties \\
    34) & Tackles & 35) &  Penalties Conceded & 36)  & Lineouts \\
    37) & Rucks & 38)  & Scrums & &\\
    \hline
  \end{tabular}
\end{table*}

% \begin{table*}[]
%   \centering
%   \caption{Features of Rugby Match Dataset}
%   \label{tab:2}
%   \begin{tabular}{c c c}
%     1) Home Team & 2) Away Team & 3) Results \\
%   \end{tabular}
% \end{table*}

\begin{table*}[htb!]
  \centering
  \caption{Features of Basketball team statistics}
  \label{tab:3}
  \begin{tabular}{|cl|cl|cl|}
  \hline
   1) &  Home/away team & 2)  & Field Goals & 3) &  Field Goals Attempted \\
	 4) &  Three-Point Shots & 5) &  Three-Point Shots Attempted & 6) &  Free Throws\\
	 7)  & Free Throws Attempted & 8) &  Offensive Rebounds & 9) &  Defensive Rebounds\\
	 10) &  Assists & 11)  & Steals & 12) &  Blocks\\
	 13)  & Turnovers & 14) &  Total Fouls & &  \\
	 \hline
  \end{tabular}
\end{table*}

The data was first normalized. When preparing the data a training-testing split was utilised. As the data-set has a temporal aspect, one cannot use future seasons to predict those preceding it~\cite{AdhikariK.2013}. The training data comprised of the first three seasons and testing data of the fourth and final season, helping to prevent any data leakage~\cite{Cerqueira2019}. However, due to limited seasons cross-validation was used for hyper-parameter tuning on the three seasons of training data. Initially, the first two seasons are used for training and validated on the third, then the first and third are used for training and validated on the second dataset. Lastly, the model is trained on the last two seasons and validated on the first.

\subsubsection{Rugby}

The data collected for rugby is from the Super Rugby International Club Competition. The data is collected from the season of 2017 to 2020. The models are trained and validated on the Super Rugby seasons of 2017, 2018, and 2019 and the season of 2020 to test the models. Data from each season is split into two parts: seasonal statistics of each team as per the end of the season; and the second being the record of every game played with the results thereof. It is noted that draws appear less than non-draws so they've been included as a home team win for creating the embeddings.

The purpose of these models is to use the team seasonal data to predict the outcome of matches between two sides for a given season. Subsequently using these predicted match outcomes to rank teams for the season. In each season there are one hundred and twenty games, with each team playing sixteen games in the regular season (each team plays eight games within its conference and eight games against teams in other conferences).

Data for the seasonal statistics are retrieved from Rugby4cast and Statbunker. The features in the datasets are presented in Table~\ref{tab:1}.

\subsubsection{Basketball}
The dataset used for basketball in this study consists of four consecutive regular seasons spanning from 2014 to 2018, each of the seasons contains 2460 games~\cite{thompson}. The data was divided into two sets, each with fourteen features. One of the sets represented the home team and the other set represented the away team. The features were chosen are as per Thabtah \textit{et al.} (2019)~\cite{article1} and Ahmadalinezhad \textit{et al.} (2019)~\cite{Ahmadalinezhad2019519},  and are shown in Table~\ref{tab:3}.

\subsection{Algorithms}
We combine gradient boosting with Siamese neural networks to improve overall performance. The reason for the combination is that the gradient boosting frameworks are effective at ranking tasks but limited in their ability to learn feature embeddings. We compare two gradient boosting frameworks combined with two Siamese network embedding losses.

XGBoost and LightGBM are Machine Learning decision-tree-based ensembles that use a gradient boosting framework for ranking, classification, and many other machine learning tasks~\cite{Ke2017}. Gradient boosting is selected over Adaptive Boosting, Random Forests and SVM's. These machines are recent extensions of those algorithms and they containing additional features aiding in performance, speed, and scalability~\cite{Rahman2020}. The result is that gradient boosting is more suitable for the desired rankings and predictions~\cite{Ke2017}.

Boosting refers to a machine learning technique that trains models iteratively~\cite{Ke2017}. A strong overall prediction is produced by starting with a weak base model which is then added to iteratively by including further weak models. Each iteration adds an additional weak model to overcome the shortcomings of the prediction of the previous weak models. The weak models are trained by applying gradient descent in the direction of the average gradient of the leaf nodes of the previous model~\cite{Ke2017}. This gradient is calculated with respect to the pseudo residual errors of the previous models loss function.

In the case of decision trees, an initial fit, $F_{0}$, and a differentiable loss function $L$ for a number of boosting rounds $M$ are used as starting points: 
\begin{equation}
F_0(x) = \underset{\gamma}{arg\ \min} \sum^{n}_{i=1} L(y_i, \gamma).
\end{equation}

Thereafter, the pseudo residuals are calculated by
\begin{equation}
  r_{im} = -\frac{\partial L(y_i, F_{m-1}(x_i))}{\partial F_{m-1}(x_i)}.
\end{equation}

Lastly, the decision tree is fit, $h_{m}(x)$ to $r_{im}$ and let
\begin{equation}
  F_m(x) = F_{m-1}(x) + \lambda_m \gamma_m h_m(x)
\end{equation}
where $ \lambda_m $ is the learning rate, $\gamma_m$ is the step size optimized by performing a line search and $y_i$ is the correct outcome of the game.

\subsubsection{LightGBM}
The first implemented evaluated is LightGBM, a gradient boost machine developed by Microsoft which uses tree-based learning algorithms. A LambdaRank~\cite{burges2010ranknet} objective function was selected to scales the gradients by the change in NDCG computed in the LightGBM~\cite{burges2010ranknet}.
\subsubsection{XGBoost}
Second XGBoost was selected an optimised distributed gradient boosting library that uses a LambdaMART objective function~\cite{burges2010ranknet}. XGBoost is designed to be highly flexible, efficient, and portable to implement machine learning algorithms under the Gradient Boosting framework. The XGBoost library has many models in it but the selected one was Pairwise ranker for these experiments. XGBoost used a Log Loss function for pairwise ranking. The Log Loss function used to calculate the cost value is given by:
\[
F = \frac{-1}{N}\sum_{i=1}^{N}[y_i \log( \hat{y}_i) + (1-y_i) \log(1-\hat{y}_i)
\]
where $y$ is the correct outcome of a game, $\hat{y}$ is the predicted result of a game and $N$ is the number of games played.

\subsubsection{Siamese Neural Network}
Data was fed into a Siamese Neural Network to create the embeddings to be passed to the aforementioned ranking algorithms. An SNN consists of two identical sub-networks where each is fed one of a selected tuple of inputs, in this case, the home and away team for each game in the data-set \cite{10.5555/2987189.2987282}. The Rectified Linear Activation Function was selected~\cite{article4} and which was used alongside an RMSprop optimizer for thirteen epochs.Thirteen epochs was selected as the model accuracy had converged  on the validation set at thirteen. As for the loss function, two were implemented: Contrastive loss and Triplet loss~\cite{Bui}. 

Contrastive loss is defined as:
\begin{equation}
  J(\theta)=(1-Y)(\frac{1}{2})(D_{a,b})^2+(Y)(\frac{1}{2})(\max(0,m-D_{a,b}))^2
\end{equation}
where Y is representative of the outcomes classes, 0 indicating a win and 1 indicating a loss, m being the dissimilarity margin and $D_{a,b}$ the euclidean distance,
\begin{equation}
  D_{a,b}(x_1,x_2)=\sqrt{\sum^{K}_{k=0}(b(x_2)_k-a(x_1)_k)^2}
\end{equation}
where $a(x_1)$ is the output from the sub-network one and $b(x_2)$ the output of the sub-network two~\cite{Bui,Hadsell2006}. A vector containing the home team's seasonal statistics is fed into the sub-network one and similarly the away team is fed into the sub-network two.

Triplet loss is defined as:
\begin{equation}
  J(\theta)=\max(D_{a,b}(a,p)-D_{a,b}(a,n)+m,0)
\end{equation}
whereby anchor (a) is an arbitrary data point, positive (p) which is the same class (win or loss) as the  anchor, and negative (n) which is a different class from the anchor~\cite{Bui}.

Each SNN is fed the seasonal data set where the respective outputs are run through the LightGBM ranker, resulting in predicted values (similarity scores) for each game in the selected seasons. To compute the final standings we use a Tally rank, if a team wins a game the similarity score is added to their tally, and likewise, if a team loses, the similarity score is subtracted from their tally.

A list-wise predicted ranking of both the Rugby and Basketball test seasons were produced. This list-wise ranking was sorted in descending order to achieve the final predicted standings. The teams were then separated into the Western and Eastern conferences for analysis purposes. From the three methodologies discussed, six results were outputted, one from each of the LightGBM and XGBoost model individually and a total of four from the LightGBM and XGBoost SNN models. Thus, a total of six predicted standings were achieved for each sport, each of which will be compared to one another using several metrics.

\subsection{Metrics}
The predicted rankings were compared to one another, to the baselines and to the actual standings of that year. The metrics that were used to evaluate the performance of the models are:

\subsubsection{Mean Average Precision (mAP)}
\begin{equation}
  mAP=\frac{1}{N}\sum^N_{i=1} AP_i
\end{equation}
with 
\begin{equation}
  AP=\sum^K_{k=1} \frac{\text{True Positives}}{\text{True Positives}+\text{False Positives}}
\end{equation}
and $K$ is the desired point where precision is calculated. The AP was evaluated at k=15 in both the Eastern and Western conferences, where the mAP is achieved by averaging these two results. Super Rugby only uses the Average Precision.

\subsubsection{Spearman's rank correlation coefficient}
\begin{equation}
  r_s=1-\frac{6\sum d_i^2}{n(n^2-1)}
\end{equation}
where $d_i$ = difference in the actual ranking of a team and the predicted rank, and $n$ the number of cases. $r_s$ can take on values between 1 and -1, where 1 indicates a perfect association of ranks, 0 indicates no association between ranks and -1 indicates a perfect negative association of ranks. The closer to zero $r_s$ is, the weaker the association between the ranks are~\cite{EllisandVictoria2011}.

\subsubsection{Normalized Discounted Cumulative Gain}
To proceed with Normalized Discounted Cumulative Gain (NDCG), one first needs to introduce relevance scores for the actual rankings of the test season under consideration. If a team finishes first in their league/conference, a score of 15 is assigned to them, the team who finishes in second place is assigned score of 14, and so forth until the fifteenth/last team receives a score of 1. These scores will be referred to as $rel$ where the respective values at position p are computed~\cite{inproceedings1}:
\begin{equation}
  nDCG_p=\frac{DCG_p}{IDCG_p}
\end{equation}
where 
\begin{equation}
  DCG_p=\sum^{p}_{i=1}\frac{2^{rel_i} -1}{\log_2(i+1)}
\end{equation}
and
\begin{equation}
  IDCG_p=\sum^{\lvert REL_P \rvert}_{i=1}\frac{2^{rel_i} -1}{\log_2(i+1)}
\end{equation}

\subsection{Experimental procedure}
First the two baselines are computed, one assigning random rankings and the other naively. For random the league standings were randomized for each sport thirty times to produce the baseline. For the naive baseline the assumption is made that the test season ranking will be identical to that of the preceding season.

Once the baselines had been established, LightGBM and XGBoost were used without the embedding from the SNN. Both algorithms are fit to the training data and rank the respective test seasons (2019/20 for rugby and 2017/18 for the NBA). A final predicted league standing was computed for both sports, the NBA is divided into the Western and Eastern conference, by adding the resulting similarity score to each team’s tally if they win, and similarly subtracting the similarity score if they lose.

Next, two SNN’s were built (only differing by their loss functions) to set up the second set of algorithms. The SNN which was built has two identical sub-networks, both of which have an input layer consisting of $50$ neurons for basketball and $37$ input neurons for rugby, followed by two hidden layers with $70$ and $20$ neurons and finally output layers with $1$ neuron respectively. Both of the sub-networks use a ReLU (Rectified Linear Activation Function)~\cite{article4} and then RMSprop optimizer for thirteen epochs. Our RMSprop optimizer had a learning rate of 0.001, a momentum value of 0, the epsilon value was $1^{-7}$ and a discounting factor for the gradient of 0.9. For contrastive loss, one of the sub-networks is fed the home team’s data and the other is fed the away team’s data. For triplet loss, a team is chosen as a baseline and the distance to the positive input is minimized and the distance to the negative input is maximized. The SNN’s were then trained on the first three seasons and were used to predict the final test season. The resulting scores were fed into both the LightGBM and XGBoost rankers from which the predicted standings were computed using the tally rank method previously discussed.

\subsection{Results and Discussion}
\begin{table}[htb!]
\caption{Basketball results}
\label{tab:5}
\centering
\resizebox{\columnwidth}{!}{%
\begin{tabular}{l|rrr}
  \toprule
  {Method} & {$mAP$} & {$r_s$} & {$NDCG$} \\ 
  \midrule \midrule
  {Naive}                  & 0.704                    & 0.640                    & 0.839           \\ 
  {Randomized}             & 0.477 $\pm0.02$          & 0.558 $\pm0.01$          & 0.820 $\pm0.05$ \\
  {LightGBM}               & 0.822 $\pm0.00$          & 0.839 $\pm0.00$          & 0.979 $\pm0.00$ \\ 
  {XGBoost}                & 0.857 $\pm0.00$          & \textbf{0.900} $\pm0.00$ & 0.976 $\pm0.00$ \\ 
  {LightGBM (Contrastive)} & 0.859 $\pm0.22$          & 0.876 $\pm0.13$          & 0.977 $\pm0.02$ \\
  {XGBoost (Contrastive)}  & 0.745 $\pm0.26$          & 0.751 $\pm0.13$          & 0.916 $\pm0.08$ \\
  {LightGBM (Triplet)}     & \textbf{0.867} $\pm0.15$ & 0.870 $\pm0.11$          & \textbf{0.980} $\pm0.00$ \\ 
  {XGBoost (Triplet)}      & 0.725 $\pm0.26$          & 0.675 $\pm0.14$          & 0.922 $\pm0.11$ \\ 
  \bottomrule
\end{tabular}}
\end{table}
 
\begin{table}[htb!]
  \centering
  \caption{Rugby results}
  \label{tab:4}
  \resizebox{\columnwidth}{!}{%
  \begin{tabular}{l|rrr}
    \toprule
    Model & {$AP$}  & {$r_s$} & {$NDCG$} \\
    \midrule \midrule
    Naive                   & 0.462                        & 0.532                      & 0.915           \\
    Randomized              & 0.193 $\pm0.20$              & 0.525 $\pm0.02$            & 0.735 $\pm0.10$ \\
    LightGBM                & 0.381 $\pm0.08$              & 0.379 $\pm0.06$            & 0.806 $\pm0.09$ \\
    XGBoost                 & 0.421 $\pm0.05$              & 0.432 $\pm0.01$            & 0.910 $\pm0.00$ \\
    LightGBM (Constrastive) & 0.620 $\pm0.02$              & 0.671 $\pm0.04$            & 0.965 $\pm0.05$ \\
    XGBoost (Contrastive)   & 0.724 $\pm0.08$              & 0.746 $\pm0.04$            & 0.974 $\pm0.01$ \\
    LightGBM (Triplet)      & 0.686 $\pm0.02$              & 0.754 $\pm0.05$            & 0.970 $\pm0.00$ \\
    XGBoost (Triplet)       & \textbf{0.921} $\pm0.04$     & \textbf{0.793} $\pm0.00$   & \textbf{0.982} $\pm0.03$\\
    \bottomrule
  \end{tabular}}
\end{table}

\subsubsection{Basketball}

All of the six models (LightGBM, LightGBM (Contrastive loss), LightGBM (Triplet loss), XGBoost, XGBoost (Contrastive loss), XGBoost (Triplet loss)) used to predict the NBA test season outperformed both the naive and random baselines in every metric (F1 score, mAP, NDCG and $r_s$). This is expected as basketball in itself is a volatile sport~\cite{inproceedings3}, making a naive or random prediction highly unlikely to resemble any form of result close to that produced by a deep similarity learning.

Arguably the most important feature of the NBA regular season is progressing to the playoffs~\cite{inproceedings3}, which only the top 8 teams from each conference advance to. The deep similarity learning models developed and investigated for this study were able to predict the playoff teams to a relatively high degree of accuracy, with the LightGBM (Triplet loss) performing the best with 8/8 Eastern Conference playoff teams correct and 7/8 Western Conference playoff teams correct. XGBoost (Triplet loss) performed the worst with only managing 6 correct from either conference. It is worth mentioning that each of the six implemented models incorrectly predicted that the Denver Nuggets would make it to the playoffs over the Utah Jazz. This is likely attributed to the Utah Jazz drafting higher quality players that year (2017/18) possibly aiding them over-perform what was expected.

Using a LightGBM without an SNN falls short in playoff predictions, F1 score, mAP, NDCG, and $r_s$, as seen in Table IV, providing us with a direct indication that the use of an SNN in unison with a LightGBM has an overall beneficial impact on the performance of the model. In terms of the selection of the loss function, our results have little differentiation between them, specifically their F1 score, NDCG, and $r_s$. However, the mAP of each loss function was the biggest discerning factor, with the Triplet loss producing a mAP of 0.867 versus the Contrastive loss mAP of 0.859. Not only does the Triplet loss improve the models' overall precision, it predicted one more correct playoff team (Washington Wizards) that the Contrastive loss.

However, there appears to be a converse effect with the use of an SNN with the XGBoost. The XGBoost function by itself outperforms the iterations where an SNN was used in every metric. From Table~\ref{tab:5}, it is clear the LightGBM (Triplet loss) algorithm produces the overall best set of results closely followed by the XGBoost algorithm.

\subsubsection{Rugby}
Looking at the scores across the models~\ref{tab:4}, the Siamese Triplet Ranker performs the best followed closely by the Siamese Contrastive, LightGBM (Triplet loss), and LightGBM (Contrastive loss). The results then fall short of our Previous Season naive implementation with the XGBoost model and LightGBM model falling behind in sixth and seventh position before the Random ranker in last.

The top-ranked amongst these models was the Siamese Triplet ranker, followed closely by the Siamese Triplet with Gradient Boosting, Siamese Contrastive and Siamese Contrastive with Gradient Boosting. The LambdaMart Gradient Boost and LambdaRank Gradient Boost models performed the worst on average and had scores that would lose out to the naive implementations.

Similarly to the NBA regular season, it's important in SuperRugby to note which are the top 8 teams move to the knockout stages. The Siamese Triplet model correctly guessed 7 out of 8 teams, whilst the rest of the trained models predicted 6 correct. The random ranker came in last with only 4 teams predicted correctly. 

\subsection{Future improvements}
Although the Triplet loss edges the performance over the Contrastive loss, the margin is slightly smaller than expected~\cite{hoffer2018deep}. The predictive models are categorized under supervised learning, which is widely accepted to be the domain of Triplet loss functions based on strategies explored in recent literature~\cite{Bui,taha2020boosting}, which indicates the under performance of the models. Triplet loss functions are sensitive to noisy data~\cite{hoffer2018deep}, which may be the hindering factor. An exploration of a new data-set, or even a new sport in itself, may help rectify this. Furthermore, an investigation into different sports would help one grasp the portability and effectiveness of the proposed predictive modeling outside of Rugby and Basketball, as it is clear that there is no one consistent best performing model across the two sports. 

\section{Conclusion}
The seasonal rankings of the National Basketball Association and Super Rugby Club Championship were predicted using Siamese Neural Network models, Gradient Boost Machines, and a combination of the two. Super Rugby is far more difficult to predict due to each team playing far fewer games compared to teams in the NBA.

In Basketball, the six models all produce satisfactory results, with the best model predicting at least 81\% of the playoff teams correctly. These models exceed a score of 0.8 in mAP, NDCG, and $r_s$ as well as outperform a majority of current literature in this field~\cite{Barlos2015,Lu2019252,article}. The models that utilise the LightGBM ranking model combined with the Siamese (Triplet Loss) Neural Network produced the highest ranking in both the NDCG and $mAP$ scores. All models surpassed both the Naive and Randomized baseline in all three metrics.

The models for rugby that utilise the Siamese (Triplet Loss) Neural Network have proven to be better than those that utilise the Siamese (Contrastive Loss) with the best-predicting 87.5\% of the play-off teams correctly and exceeding 0.9 in mAP, NDCG, and $r_s$. The models that use LightGBM and XGBoost are not as effective as their counterparts which do not. Both the LightGBM and XGBoost models which do not use the SNN as part of their design failed to provide better results than the Naive rankings but were better than the Randomized list.

Using Gradient Boost Machines and SNNs, it is possible to create impressive and capable ranking and prediction models for the NBA and Super Rugby. However, there is no one best performing model across the two sports. This research can be built upon to further delve into more sporting disciplines or expand on the current models by training on more seasons and better feature modelling. It may be concluded that a LightGBM model is more effective in ranking larger data-sets, as prevalent in the NBA, compared to the XGBoost model which explains the differences in their respective results.

In conclusion, the models proposed provided excellent results in comparison to their baseline counterparts and answered the research questions at hand. They have been proven that they can be utilized in betting industries to forecast results that are precise in both the Super Rugby and NBA.
\bibliography{References}

\begin{thebibliography}{10}
\providecommand{\url}[1]{#1}
\csname url@rmstyle\endcsname
\providecommand{\newblock}{\relax}
\providecommand{\bibinfo}[2]{#2}
\providecommand\BIBentrySTDinterwordspacing{\spaceskip=0pt\relax}
\providecommand\BIBentryALTinterwordstretchfactor{4}
\providecommand\BIBentryALTinterwordspacing{\spaceskip=\fontdimen2\font plus
\BIBentryALTinterwordstretchfactor\fontdimen3\font minus
  \fontdimen4\font\relax}
\providecommand\BIBforeignlanguage[2]{{%
\expandafter\ifx\csname l@#1\endcsname\relax
\typeout{** WARNING: IEEEtran.bst: No hyphenation pattern has been}%
\typeout{** loaded for the language `#1'. Using the pattern for}%
\typeout{** the default language instead.}%
\else
\language=\csname l@#1\endcsname
\fi
#2}}

\bibitem{AdhikariK.2013}
R.~{Adhikari K.} and A.~R.K., ``{An Introductory Study on Time Series Modeling
  and Forecasting Ratnadip Adhikari R. K. Agrawal},'' \emph{arXiv preprint
  arXiv:1302.6613}, 2013.


\bibitem{article4}
A.~F. Agarap, ``Deep learning using rectified linear units (relu),''
  \emph{arXiv preprint arXiv:1803.08375}, 03 2018.


\bibitem{Ahmadalinezhad2019519}
M.~Ahmadalinezhad, M.~Makrehchi, and N.~Seward, ``Basketball lineup performance
  prediction using network analysis,'' in \emph{Basketball lineup performance
  prediction using network analysis}, 2019, pp. 519--524, cited By 0.


\bibitem{Arnold2012}
T.~M. Arnold and J.~M. Godbey, ``{Introducing Linear Regression: An Example
  Using Basketball Statistics},'' \emph{SSRN Electronic Journal}, vol.~11,
  no.~2, pp. 113--130, 2012.


\bibitem{Barlos2015}
K.~Barlos and S.~Koutsogianni, ``{Predicting Regular Season Results of NBA
  TeamsBased on Regression Analysis of Common Basketball Statistics},''
  \emph{UNIVERSITY OF CALIFORNIA AT BERKELEY}, vol.~97, no.~12, pp. 194--200,
  2015.


\bibitem{10.1145/130385.130401}
\BIBentryALTinterwordspacing
B.~E. Boser, I.~M. Guyon, and V.~N. Vapnik, ``A training algorithm for optimal
  margin classifiers,'' in \emph{Proceedings of the Fifth Annual Workshop on
  Computational Learning Theory}, ser. COLT ’92.\hskip 1em plus 0.5em minus
  0.4em\relax New York, NY, USA: Association for Computing Machinery, 1992, p.
  144–152. [Online]. Available: \url{https://doi.org/10.1145/130385.130401}
\BIBentrySTDinterwordspacing


\bibitem{10.5555/2987189.2987282}
J.~Bromley, I.~Guyon, Y.~LeCun, E.~S\"{a}ckinger, and R.~Shah, ``Signature
  verification using a “siamese” time delay neural network,'' in
  \emph{Proceedings of the 6th International Conference on Neural Information
  Processing Systems}, ser. NIPS’93.\hskip 1em plus 0.5em minus 0.4em\relax
  San Francisco, CA, USA: Morgan Kaufmann Publishers Inc., 1993, p. 737–744.


\bibitem{Bui}
T.~Bui, J.~Collomosse, L.~Ribeiro, T.~Nazare, and M.~Ponti, ``{Regression in
  Deep Learning: Siamese and Triplet Networks},'' \emph{2017 30th SIBGRAPI
  conference on graphics, patterns and images tutorials (SIBGRAPI-T)}, no.
  Icmc, 2017.


\bibitem{burges2010ranknet}
C.~J. Burges, ``From ranknet to lambdarank to lambdamart: An overview,''
  \emph{Learning}, vol.~11, no. 23-581, p.~81, 2010.


\bibitem{Cerqueira2019}
\BIBentryALTinterwordspacing
V.~Cerqueira, L.~Torgo, and I.~Mozetic, ``{Evaluating time series forecasting
  models: An empirical study on performance estimation methods},''
  \emph{Machine Learning}, pp. 1--28, 2019. [Online]. Available:
  \url{http://arxiv.org/abs/1905.11744}
\BIBentrySTDinterwordspacing


\bibitem{inproceedings}
Y.~Chen, J.~Dai, and C.~Zhang, ``A neural network model of the nba most valued
  player selection prediction,'' in \emph{A Neural Network Model of the NBA
  Most Valued Player Selection Prediction}, 08 2019, pp. 16--20.


\bibitem{article}
G.~Cheng, Z.~Zhang, M.~Kyebambe, and K.~Nasser, ``{Predicting the Outcome of
  NBA Playoffs Based on the Maximum Entropy Principle},'' \emph{Entropy},
  vol.~18, p. 450, 2016.


\bibitem{dlamini2019author}
N.~Dlamini and T.~L. van Zyl, ``Author identification from handwritten
  characters using siamese cnn,'' in \emph{2019 International Multidisciplinary
  Information Technology and Engineering Conference (IMITEC)}.\hskip 1em plus
  0.5em minus 0.4em\relax IEEE, 2019, pp. 1--6.


\bibitem{EllisandVictoria2011}
{Ellis and Victoria}, ``{Spearma n ' s correlation},'' \emph{Ellis and Victoria
  textbook}, p.~8, 2011.


\bibitem{Goodfellow-et-al-2016}
I.~Goodfellow, Y.~Bengio, and A.~Courville, \emph{Deep Learning}.\hskip 1em
  plus 0.5em minus 0.4em\relax MIT Press, 2016,
  \url{http://www.deeplearningbook.org}.


\bibitem{Hadsell2006}
R.~Hadsell, S.~Chopra, and Y.~LeCun, ``{Dimensionality reduction by learning an
  invariant mapping},'' \emph{Proceedings of the IEEE Computer Society
  Conference on Computer Vision and Pattern Recognition}, vol.~2, pp.
  1735--1742, 2006.


\bibitem{hoffer2018deep}
E.~Hoffer and N.~Ailon, ``Deep metric learning using triplet network,'' 2018.


\bibitem{Ke2017}
G.~Ke, Q.~Meng, T.~Finley, T.~Wang, W.~Chen, W.~Ma, Q.~Ye, and T.~Y. Liu,
  ``{LightGBM: A highly efficient gradient boosting decision tree},''
  \emph{Advances in Neural Information Processing Systems}, vol. 2017-Decem,
  no. Nips, pp. 3147--3155, 2017.


\bibitem{Lieder2018}
N.~Lieder, ``{Can Machine-Learning Methods Predict the Outcome of an NBA
  Game?}'' \emph{SSRN Electronic Journal}, pp. 1--10, 2018.


\bibitem{Lin2014}
J.~Lin, L.~Short, and V.~Sundaresan, ``{Predicting National Basketball
  Association Winners},'' \emph{CS 229 FINAL PROJECT}, pp. 1--5, 2014.


\bibitem{Lu2019252}
J.~Lu, Y.~Chen, and Y.~Zhu, ``Prediction of future nba games' point difference:
  A statistical modeling approach,'' in \emph{Prediction of future NBA Games'
  point difference: A statistical modeling approach}, 2019, pp. 252--256, cited
  By 0.


\bibitem{manack2020deep}
H.~Manack and T.~L. Van~Zyl, ``Deep similarity learning for soccer team
  ranking,'' in \emph{2020 IEEE 23rd International Conference on Information
  Fusion (FUSION)}.\hskip 1em plus 0.5em minus 0.4em\relax IEEE, 2020, pp.
  1--7.


\bibitem{inproceedings3}
B.~Pagno, L.~Guedes, C.~Maria, and L.~Nedel, ``Nbavis: Visualizing national
  basketball association information,'' in \emph{NBAVis: Visualizing National
  Basketball Association Information}, 08 2014.


\bibitem{Rahman2020}
S.~Rahman, M.~Irfan, M.~Raza, K.~M. Ghori, S.~Yaqoob, and M.~Awais,
  ``{Performance analysis of boosting classifiers in recognizing activities of
  daily living},'' \emph{International Journal of Environmental Research and
  Public Health}, vol.~17, no.~3, 2020.


\bibitem{SuperRugRules}
S.~RugRules, \emph{Super Rugby Rules - Super Rugby | Super 15 Rugby And Rugby
  Championship News,Results And Fixtures From Super XV Rugby}, 2016 (accessed
  November 15, 2020), \url{https://www.superxv.com/super-rugby}.


\bibitem{taha2020boosting}
A.~Taha, Y.-T. Chen, T.~Misu, A.~Shrivastava, and L.~Davis, ``Boosting standard
  classification architectures through a ranking regularizer,'' 2020.


\bibitem{article1}
F.~Thabtah, L.~Zhang, and N.~Abdelhamid, ``Nba game result prediction using
  feature analysis and machine learning,'' \emph{Annals of Data Science}, 01
  2019.


\bibitem{thompson}
A.~{Thompson}, ``\BIBforeignlanguage{English}{All the news: 143,000 articles
  from 15 american publications},''
  \url=https://www.kaggle.com/snapcrack/all-the-news, 8 2017.


\bibitem{Torres2013}
R.~A. Torres and Y.~H. Hu, ``{Prediction of NBA games based on Machine Learning
  Methods},'' \emph{University of Wisconsin, Madison}, 2013.


\bibitem{inproceedings1}
H.~Valizadegan, R.~Jin, R.~Zhang, and J.~Mao, ``Learning to rank by optimizing
  ndcg measure,'' in \emph{Learning to Rank by Optimizing NDCG Measure}, 01
  2009, pp. 1883--1891.


\bibitem{van2020unique}
T.~L. Van~Zyl, M.~Woolway, and B.~Engelbrecht, ``Unique animal identification
  using deep transfer learning for data fusion in siamese networks,'' in
  \emph{2020 IEEE 23rd International Conference on Information Fusion
  (FUSION)}.\hskip 1em plus 0.5em minus 0.4em\relax IEEE, 2020, pp. 1--6.


\end{thebibliography}
\end{document}